\documentclass{article}

\PassOptionsToPackage{numbers, compress}{natbib}

\usepackage[preprint]{neurips_2024}

\usepackage[utf8]{inputenc} 
\usepackage[T1]{fontenc}    
\usepackage{hyperref}       
\usepackage{url}            
\usepackage{booktabs}       
\usepackage{amsfonts}       
\usepackage{nicefrac}       
\usepackage{microtype}      
\usepackage{xcolor}         

\usepackage{inconsolata}
\usepackage{booktabs} 
\usepackage{pifont}
\usepackage{tikz}
\usepackage{graphicx} 
\usepackage{multirow}
\usepackage{multicol}
\usepackage{arydshln}
\usepackage{color}
\usepackage{amssymb}
\usepackage{amsmath}

\title{Language Model Can Listen While Speaking}

\author{
Ziyang Ma\textsuperscript{1,2}\quad
Yakun Song\textsuperscript{1,2}\quad
Chenpeng Du\textsuperscript{2}\quad
Jian Cong\textsuperscript{2}\quad 
Zhuo Chen\textsuperscript{2}\quad \\
\textbf{
Yuping Wang\textsuperscript{2}\quad
Yuxuan Wang\textsuperscript{2}\quad
Xie Chen\textsuperscript{1}\thanks{Corresponding author.}
}\\
\textsuperscript{1}MoE Key Lab of Artificial Intelligence, X-LANCE Lab, Shanghai Jiao Tong University \\  
\textsuperscript{2} ByteDance Inc. 
}

\begin{document}

\maketitle

\begin{abstract}
Dialogue serves as the most natural manner of human-computer interaction (HCI). 
Recent advancements in speech language models (SLM), have significantly enhanced speech-based conversational AI. 
However, these models are limited to turn-based conversation, lacking the ability to interact with humans in real-time spoken scenarios, for example, being interrupted when the generated content is not satisfactory. 
To address these limitations, we explore full duplex modeling (FDM) in interactive speech language models (iSLM), focusing on enhancing real-time interaction and, more explicitly, exploring the quintessential ability of interruption. 
We introduce a novel model design, namely listening-while-speaking language model (LSLM), an end-to-end system equipped with both listening and speaking channels. 
Our LSLM employs a token-based decoder-only TTS for speech generation and a streaming self-supervised learning (SSL) encoder for real-time audio input. 
LSLM fuses both channels for autoregressive generation and detects turn-taking in real time. 
Three fusion strategies—early fusion, middle fusion, and late fusion—are explored, with middle fusion achieving an optimal balance between speech generation and real-time interaction. 
Two experimental settings, command-based FDM and voice-based FDM, demonstrate LSLM's robustness to noise and sensitivity to diverse instructions. 
Our results highlight LSLM's capability to achieve duplex communication with minimal impact on existing systems. 
This study aims to advance the development of interactive speech dialogue systems, enhancing their applicability in real-world contexts\footnote{Demo can be found at \url{https://ddlbojack.github.io/LSLM}}.
\end{abstract}


\section{Introduction}
Dialogue is the most natural way of human-computer interaction (HCI). 
With the rapid development of GPT-style~\citep{radford2018improving} large language models (LLM) and the scaling of Transformer-style~\citep{vaswani2017attention} architectures, textual conversational AI, such as ChatGPT~\citep{ouyang2022training, achiam2023gpt} and LLaMA~\citep{touvron2023llama1, touvron2023llama2}, have become a significant part of daily life. 
However, these models are limited to text input and output and cannot interact directly with humans in arbitrary scenarios.

Incorporating spoken and auditory interfaces into conversational AI enhances HCI convenience. 
Leveraging techniques from text LLMs, the speech language model (SLM) processes speech similarly to text. 
This paradigm involves encoding the speech signal into discrete tokens or continuous embeddings, modeling them with a language model, and decoding the speech tokens or embeddings back to the speech signal. Some studies~\citep{gslm, pgslm, dgslm} utilizes this paradigm for speech continuation, generating expressive speech and natural multi-round dialogue. 
Other research employs this paradigm to task-specific applications, such as decoder-only high-fidelity TTS~\citep{wang2023neural, borsos2023audiolm, anastassiou2024seed, du2024cosyvoice} and decoder-only streaming ASR~\citep{seide2024speech, tsunoo2024decoder, chen2024streaming, chen2024bestow}
Moreover, SpeechGPT~\citep{zhang2023speechgpt} and LauraGPT~\citep{chen2023lauragpt} initialize SLMs using LLMs, expanding speech tokens to the LLM vocabulary and continuing training on speech. This empowers SLM to comprehend semantic information and equips SLM with dialogue capability. 
Despite these advances, all these models are limited to turn-based conversations and cannot handle real-time sound or interruptions, limiting their applicability in real-life scenarios. 

Interaction and turn-taking are essential abilities for natural communication among humans. 
At the dawn of the end-to-end speech dialogue system explosion, we focus on investigating \textbf{F}ull \textbf{D}uplex \textbf{M}odeling (\textbf{FDM}) in \textbf{i}nteractive \textbf{S}peech \textbf{L}anguage \textbf{M}odels (\textbf{iSLM}),  a crucial topic affecting user experience. 
Lin et. al~\cite{lin2022duplex} proposes to process real-time audio input with a separate comprehension module. 
Other works~\citep{zhang2024beyond, wang2024full} suggest modifying the order in which text tokens are organized in the LLM to tackle the duplex modeling problem. 
All these models are based on text-centric LLMs that require external ASR and TTS modules for spoken dialogue. 
As a result, latency remains perceivable and the paralinguistic ability is still lacking. 
We believe the FDM capability should be an intrinsic capability of SLMs, enabling simultaneous listening and speaking. 

To engage FDM capability for iSLM, we propose \textbf{L}istening-while-\textbf{S}peaking \textbf{L}anguage \textbf{M}odel (LSLM), an end-to-end model with both listening and speaking channels. 
The proposed LSLM uses a token-based decoder-only TTS to model the ability to speak and a streaming self-supervised learning (SSL) encoder to model the ability to listen. 
LSLM fuses these two channels and detects turn-taking in real time.
We explore three strategies for fusing duplex signals: \textbf{Early Fusion}, \textbf{Middle Fusion}, and \textbf{Late Fusion}. 
Experiments demonstrate that middle fusion achieves a good balance between speech generation and real-time interaction capabilities.

In addition, interactive dialogue systems for realistic scenarios have two important features: 
\textbf{1) Listening channels are not always clean.} Users may interact with iSLMs in different scenarios, containing high-frequency noise (e.g., telephone ringing) and low-frequency noise (e.g., white noise). 
\textbf{2) It is possible that the iSLM interacts with an unseen speaker.} iSLMs should recognize and respond to new voices and instructions, not dismiss them as noise. 
Therefore, iSLM should have both robustness to noise and sensitivity to unseen speakers. 
To test LSLM, we designed two scenarios: 
\textbf{Command-based FDM}, where LSLM is interrupted by a specific command, and \textbf{Voice-based FDM}, where LSLM can be interrupted by various words from unseen speakers. 
Experimental results show that LSLM with a listening channel is robust to noisy input and sensitive to turning-taking.

Our contributions are summarized as follows:
\begin{enumerate}
    \item We formulate an important task, \textbf{F}ull \textbf{D}uplex \textbf{M}odeling (\textbf{FDM}), applied in the interactive speech language model (\textbf{iSLM}). 
    \item We propose \textbf{L}istening-while-\textbf{S}peaking \textbf{L}anguage \textbf{M}odel (LSLM), an end-to-end single model with the focus of modeling the turn-taking problem. LSLM can listen to the outside signal and provide feedback in real time while speaking.  
    \item We introduce three methods for fusing duplex signals: \textbf{Early Fusion}, \textbf{Middle Fusion}, and \textbf{Late Fusion}, with Middle Fusion providing the optimal tradeoff between speech generation and real-time interaction. 
    \item We tested the FDM ability of the proposed LSLM in two scenarios: \textbf{Command-based FDM} and \textbf{Voice-based FDM}. Experiments indicate that our proposed LSLM can achieve duplexing capability with little impact on the previous system. 
\end{enumerate}

\begin{figure}[htbp]
  \centering
  \includegraphics[width=0.9\textwidth]{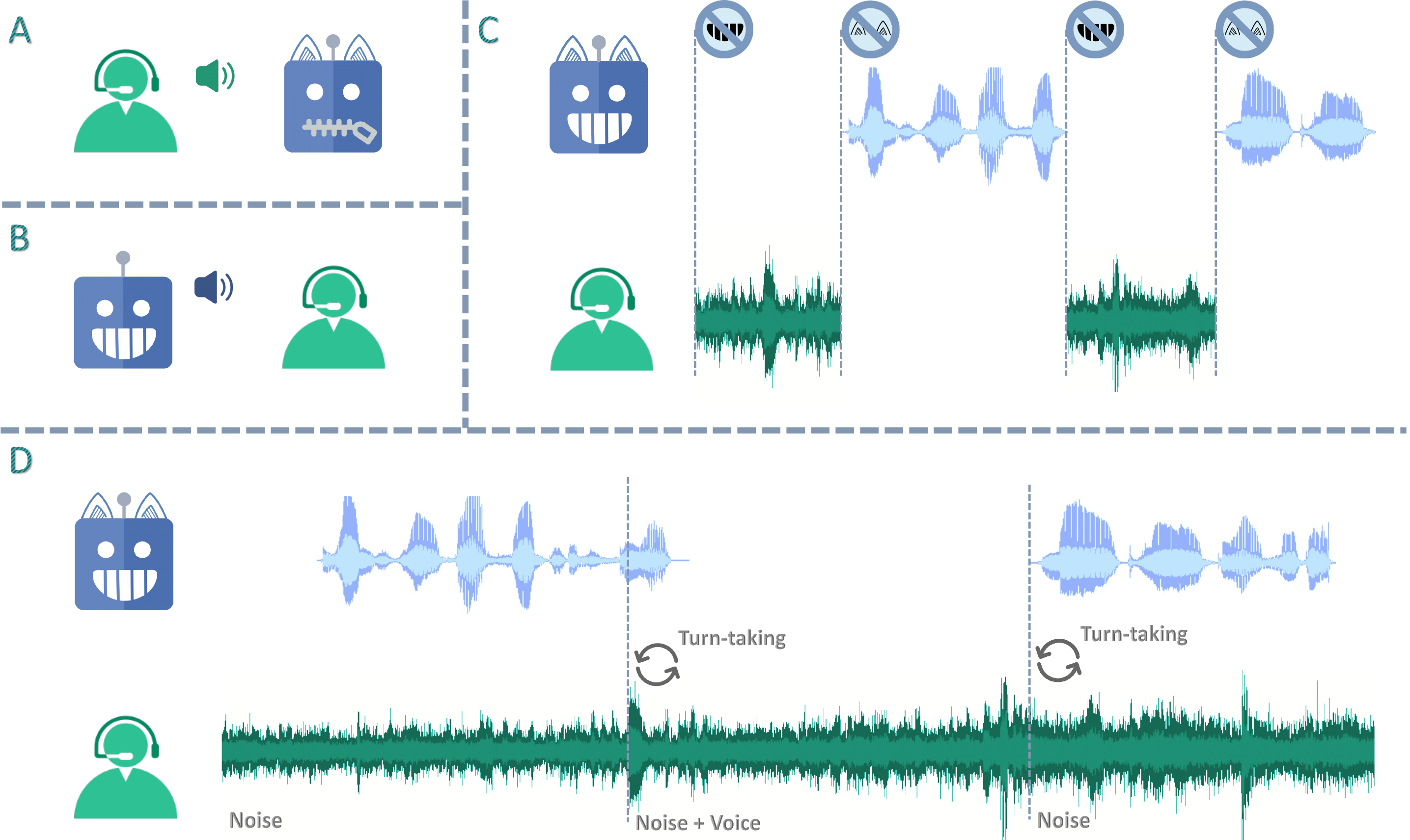}
  \caption{Illustration of simplex, half duplex, and full duplex speech language models.  (A): Simplex speech language model with listening ability. (B): Simplex speech language model with speaking ability. (C): Half duplex speech language model with both listening and speaking abilities. (D): Full duplex speech language model can listen while speaking. }
  \label{fig:duplex}
\end{figure}

\section{Related Work}

Figure~\ref{fig:duplex} illustrates the distinctions between simplex, half duplex, and full duplex speech language models from a telecommunication perspective. An SLM with full duplex modeling (FDM) capability can be referred to as an interactive speech language model (iSLM). 

\subsection{Simplex and Half Duplex Speech Language Model}
Simplex SLMs, depicted in Figure~\ref{fig:duplex}(A) and \ref{fig:duplex}(B), are limited to a single channel, either for listening or speaking. 
With the assistance of LLM, simplex SLMs exhibit strong understanding capabilities. 
Representative works include LLM-based ASR~\citep{yu2023connecting, ma2024embarrassingly, yang2024mala, bai2024seed}, LLM-based speech translation~\citep{pan2023cosmic, chen2024salm, huang2024investigating, chen2024llast}, and LLM-based speech emotion understanding~\citep{xu2024secap, lin2024advancing, lian2024affectgpt}. 
Similarly, simplex SLMs have demonstrated robust generation capabilities, as seen in LLM-based TTS~\citep{hao2023boosting, neekhara2024improving, lajszczak2024base, anastassiou2024seed}. 
Some research leverages the powerful in-context learning capabilities of LLMs to extend task-specific abilities to more universal applications, such as speech understanding~\citep{deng2024wav2prompt}, audio understanding~\citep{gong2023listen}, or both~\citep{tang2023salmonn, chu2023qwen, chu2024qwen2}. 
Despite their growing power and versatility, simplex SLMs are limited to one-way communication (either human $\rightarrow$ machine or machine $\rightarrow$ human). 
LLMs have facilitated a paradigm shift from simplex models to half-duplex models, also known as turn-based models, as shown in Figure~\ref{fig:duplex}(C).  
Prominent models include SpeechGPT~\citep{zhang2023speechgpt}, LauraGPT~\citep{chen2023lauragpt}, and VioLA~\citep{wang2023viola}. 
While these half duplex models can both listen and speak, they are constrained to performing only one action at the same instant, thus failing to address the turn-taking problem. 

\subsection{Full Duplex Speech Language Model}
Full duplex SLMs, as shown in Figure~\ref{fig:duplex}(D), have the capability to listen and speak simultaneously, allowing for turn-taking whenever a human interrupts the machine. 
Recent efforts~\citep{zhang2024beyond, wang2024full} have attempted to build full duplex capabilities on text-centric LLMs with cascade ASR and TTS modules. 
Cutting-edge products like GPT-4o~\footnote{\url{https://openai.com/index/hello-gpt-4o}} and Moshi~\footnote{\url{https://moshi.chat}} exhibit full duplex capability in their spoken dialogue systems. 
Despite these advancements, there are no publicly available open-source models or detailed analyses of full duplex SLMs. This gap highlights the need for further research and development to fully understand and optimize full duplex capability in speech language models. 

\section{Full Duplex Modeling (FDM)}
A simplex or half duplex spoken dialogue system can be modeled by finding the parameters $\theta$ that maximize the log-likelihood function, formulated as: 
\begin{equation}
\max\limits_{\theta} \sum_{(C,R)\in D} \log P_{\theta}(R|C),
\end{equation}
where $(C, R)$ represents the context-response pairs in the dataset $D$ and $P_{\theta}(R|C)$ is the probability of the response $R$ given the context $C$ and parameters $\theta$. 
More specifically, if the spoken dialogue system is modeled by an autoregressive language model where the response $R$ is generated token by token, the training loss $\mathcal{L}(\theta)$ for each sample is expressed as:
\begin{equation}
\mathcal{L}(\theta) = - \sum_{t=1}^{T} \log P_{\theta}(r_t|R_{1:t-1}, C),
\end{equation}
where $R_{1:t-1} = [r_1, r_2, ..., r_{t-1}]$ and $T$ is the sequence length. 
During the inference phase, the model can only predict the next token autoregressively based on the previous output within the current channel, without information from other channels. 

In modeling a full duplex spoken dialogue system within an autoregressive language model, the model needs to predict the next token $r_t$ in the response $R$ not only based on the context $C$ and the generated response history $R_{1:t-1} = [r_1, r_2, \ldots, r_{t-1}]$ in the current channel, but also by utilizing information $S_{1:t-1} = [s_1, s_2, \ldots, s_{t-1}]$ from another channel simultaneously. 
Here we extend the modeling approach used for simplex or half duplex dialogue systems to accommodate the requirements of full duplex modeling (FDM). The training loss $\mathcal{L}(\theta)$ is now formulated as: 
\begin{equation}
\mathcal{L}(\theta) = - \sum_{t=1}^{T} \log P_{\theta}(r_t|R_{1:t-1}, S_{1:t-1}, C)
\end{equation}
\textbf{A key point in FDM is that the sequence $S$ is produced in real time and unpredictably. }
Taking the full duplex speech language model as an example, at the inference step $t-1$, the current speaking channel generates output $r_{t-1}$ and listening channel acquired input $s_{t-1}$ are fed into the model simultaneously, influencing the prediction of the speaking channel's next step output $r_t$. 
This modeling approach endows the system with a full duplex ability, enabling it to effectively leverage the multi-channel information during dialogue, thereby improving the accuracy and fluency of the real-time interaction capability. 

\section{Proposed LSLM}

The core difference between LSLM and previous speech language models lies in its capability to simultaneously speak and listen. 
We first introduce the speaking capability of LSLM, followed by its listening capability, and finally, we discuss various fusion methods that integrate these capabilities, endowing LSLM with full duplex ability.

\begin{figure*}[htbp]
  \centering
  \includegraphics[width=0.8\textwidth]{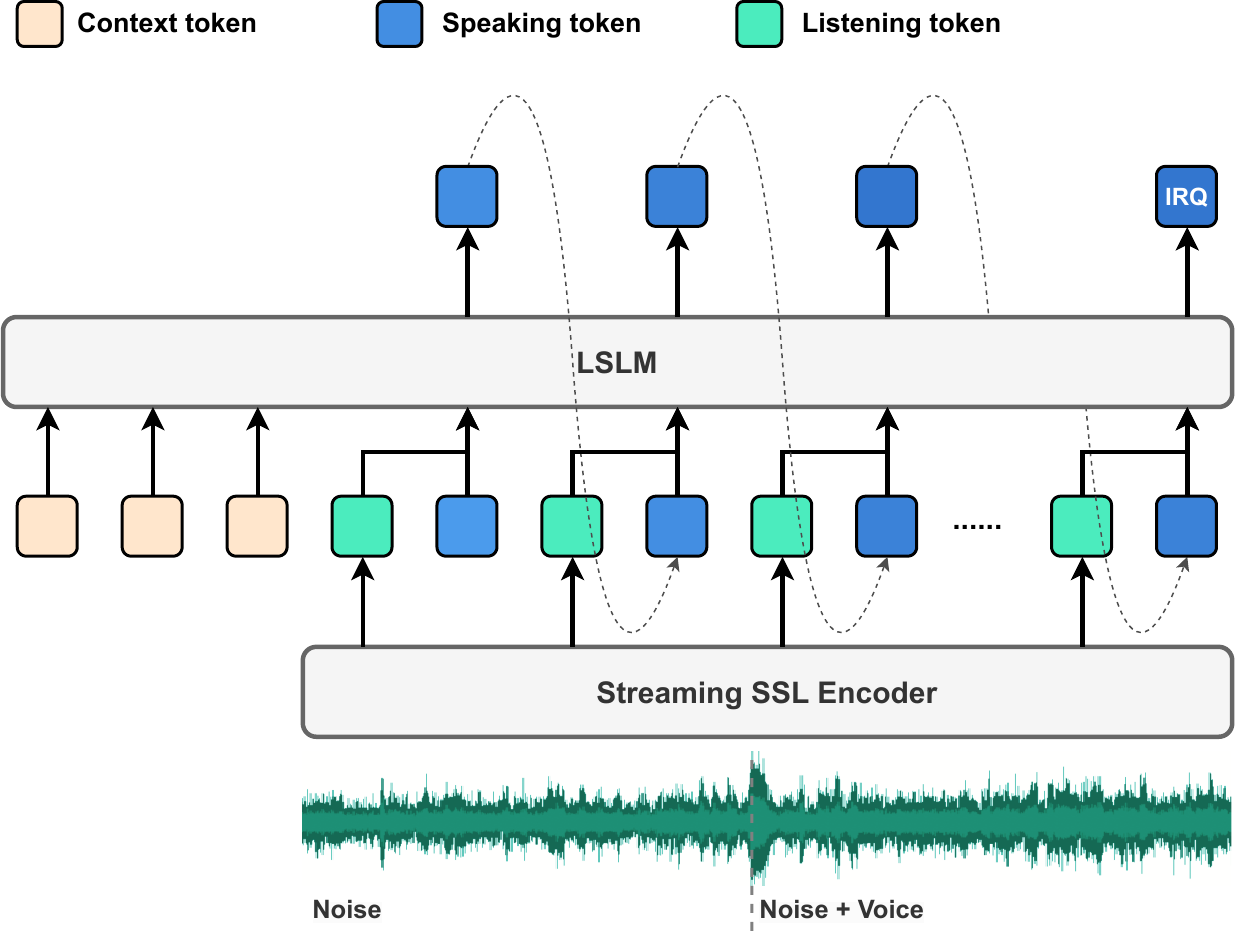}
  \caption{Proposed LSLM. The model contains a decoder-only Transformer to generate speaking tokens and a streaming SSL encoder to process listening tokens. An interruption token (IRQ) is added to allow the model to terminate early if a turn-taking occurs. }
  \label{fig:model}
\end{figure*}

\subsection{Speaking Ability}
To simulate the speaking ability of the LSLM, we utilize an autoregressive token-based TTS model. 
Unlike VALL-E-styled models that combine autoregressive (AR) and non-autoregressive (NAR) approaches with multi-layer residual vector quantization (RVQ) tokens, our model employs a single layer of discrete audio tokens. 
This design better meets the requirements for real-time interaction, as it eliminates the need to wait for the completion of AR token synthesis before performing NAR operations. 
Given target speech $X^R$, an SSL encoder $Enc$ is utilized to obtain a continuous embedding $R$, which can be written as: 
\begin{equation}
\label{eq:ssl_encoder}
R = Enc(X^R).
\end{equation}

To train an autoregressive TTS model based on discrete tokens, we quantize the speech embedding $R$, denoted by: 
\begin{equation}
R^q = Qnt(R),
\end{equation}
where $Qnt$ is the discretization operation and $R^q$ are the discrete tokens. 
Given the context information $C$, in this scenario the text content to be synthesized, the model synthesizes the corresponding speech discrete tokens autoregressively. We minimize the negative log-likelihood of the target sequence to train the decoder-only model, conditioned on the preceding tokens and the context. The loss function is defined as:
\begin{equation}
\label{eq:speaking_ability_loss}
\mathcal{L}(\theta_{S}) = - \sum_{t=1}^{t_{EOS}} \log P(r^q_t|R^q_{1:t-1}, C; \theta_{S}),
\end{equation}
where $\theta_{S}$ are the parameters to model speaking ability, $t_{EOS}$ represents the time step at which the end-of-sequence token is reached,  $r^q_t$ is the target discrete token at time step $t$, $R^q_{1:t-1}$ denotes the sequence of all previous tokens up to time step $t-1$, and $C$ is the text content to be synthesized. During inference, the model samples $\hat{r}^q_t$ from a conditional distribution based on the already generated tokens $\hat{R}^q_{1:t-1}$ and the context $C$. The process is described by the following equation: 
\begin{equation}
 \hat{r}^q_t\sim P(r^q_t | \hat{R}^q_{1:t-1}, C; \theta_{S}).
\end{equation}

A vocoder $Dec$ is employed to recover the speech signal $\hat{X}^R$ from discrete tokens $\hat{R}^q$, donated by: 
\begin{equation}
\label{eq:speaking_ability_inference}
\hat{X}^R = Dec(\hat{R}^q, A),
\end{equation}
where $A$ is the acoustic prompt providing the timbre of the synthesized speech. 
This decoupling of timbre from content allows the AR model to focus more on semantic information rather than paralinguistic information.

\subsection{Listening Ability}
Given the audio input $X^S$ of the listening channel, the same SSL encoder $Enc$ in Equation~\ref{eq:ssl_encoder} is used to obtain a continuous embedding $S$, which can be written as:
\begin{equation}
S = Enc(X^S),
\end{equation}
where $X^S$ can be a variety of sound signals, including environmental noise and human speech. Unlike training the speaking ability, which involves a discretization module, the listening channel embedding $S$ is fed into the neural network end-to-end via a projection module $Proj$, which can be written as:
\begin{equation}
S^p = Proj(S),
\end{equation}
where the listened audio signal is mapped to a space that can be processed by the AR model. 

\subsection{FDM Ability}
\label{sec:FDM-Ability}
LSLM has two channels: speaking and listening. At time step $t$, all previous information of the speaking channel $R^q_{1:t-1}$ and the processed information of the listening channel $S^p_{1:t-1}$ are considered by the model simultaneously. Here we revise Equation~\ref{eq:speaking_ability_loss} as follows:
\begin{equation}
\mathcal{L}(\theta_{LS}) =
\begin{cases} 
- \sum_{t=1}^{t_{IRQ}} \log P(r^q_t|R^q_{1:t-1}, S^p_{1:t-1}, C; \theta_{LS}) & \text{if turn-taking,} \\
- \sum_{t=1}^{t_{EOS}} \log P(r^q_t|R^q_{1:t-1}, S^p_{1:t-1}, C; \theta_{LS}) & \text{otherwise.} \\
\end{cases}
\end{equation}
where $\theta_{LS}$ are the parameters to model the proposed LSLM with listening-while-speaking ability. 
In addition to the EOS token, we add an interruption token IRQ to the tokenizer vocabulary to allow the model to terminate early if turn-taking occurs. For example, if a human interrupts, the model should stop speaking within a detection interval $\mu$ seconds after the interruption starts. 
During inference, the model samples $\hat{r}^q_t$ from a conditional distribution based on the already generated tokens $\hat{R}^q_{1:t-1}$, the context $C$, and most important, real-time listened audio tokens $S^p_{1:t-1}$. The revised formula from Equation~\ref{eq:speaking_ability_inference} is written as follows:
\begin{equation}
 \hat{r}^q_t\sim P(r^q_t | \hat{R}^q_{1:t-1}, S^p_{1:t-1}, C; \theta_{LS}),
\end{equation}
in which, an essential requirement for the SSL encoder $Enc$ is that it is streaming. Thus, LSLM can obtain real-time audio features during inference. This is detailed further in Section~\ref{sec:model_details}.

\begin{figure*}[htbp]
  \centering
  \includegraphics[width=1\textwidth]{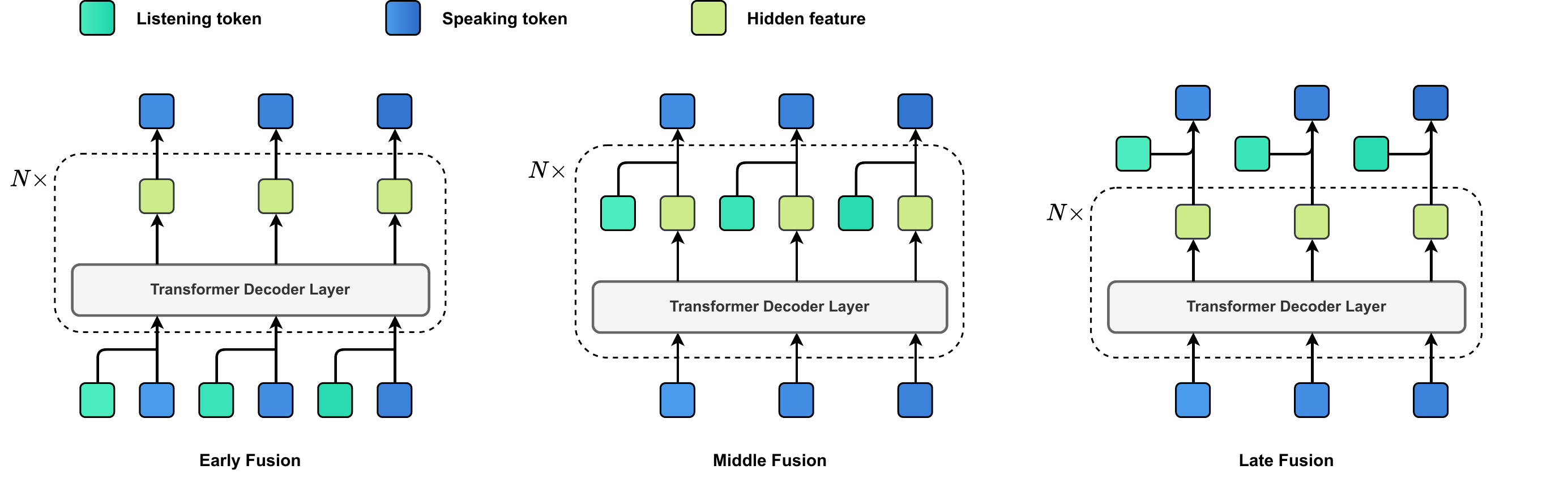}
  \caption{Different model designs to integrate the listening channel to the proposed LSLM. }
  \label{fig:fusion}
\end{figure*}

To comprehensively explore the integration of a listening channel to the proposed LSLM, we try to fuse the listening channel and the speaking channel with early, middle, and late methods, as shown in Figure~\ref{fig:fusion}. 
\paragraph{Early Fusion} integrates the listening and speaking channels at the input embeddings before autoregressive prediction. 
\paragraph{Middle Fusion} merges the listening and speaking channels at each Transformer block. Specifically, in addition to the hidden states of the speaking channel and positional embeddings, the listening channel is additionally added to the input of each Transformer block. 
\paragraph{Late Fusion} combines the channels at the output logits before the softmax operation. 

\section{Setup}
\subsection{Model Details}
\label{sec:model_details}
The backbone of the proposed LSLM employs a decoder-only Transformer architecture consisting of $12$ Transformer blocks, $12$ attention heads, $768$ embedding dimensions, and $3072$ feed-forward layer dimensions, resulting in $106$M parameters. 
SSL encoder vq-wav2vec~\citep{baevskivq} is employed to extract audio features and further convert speech features to discrete tokens. 
vq-wav2vec, a fully convolutional self-supervised pre-trained model with $20$ layers of 1D convolutional neural networks with $34$M parameters, is naturally suitable for streaming audio feature extraction. 
A simple linear layer serves as the projection module to adapt the listening channel features to the AR model. 
A GAN-based token-to-waveform vocoder~\citep{du2024unicats} is utilized to recover discrete audio tokens to speech waveform.

\subsection{Data Details}
We evaluate the proposed LSLM under two full duplex modeling (FDM) settings: command-based FDM and voice-based FDM. 
Table \ref{tab:data_details} summarizes the datasets and experimental settings. 
For the TTS datasets, we utilize the LibriTTS dataset~\citep{zen2019libritts} with $585$ hours of speech-text pairs for training and validation. LibriTTS-testsetB~\citep{du2024unicats} is adopted for testing, which contains $500$ utterances sampled from the test-clean subset of LibriTTS with 37 unseen speakers. 
Background noise is uniformly sourced from the Freesound portion of the MUSAN dataset~\citep{snyder2015musan}, which includes high-frequency noise such as telephone ringing and sounds of the explosion, as well as low-frequency noise such as white noise and traffic noise. 
The model needs to distinguish the human voice from the noise, so as to avoid turning-taking with any random input signals and avoid trivial solutions. 
Different interruption data is constructed based on the FDM settings.
\paragraph{Command-based FDM. }In this setting, LSLM can only be interrupted by specific keywords. Timbre of $22$ boutique speakers from SEED-TTS~\cite{anastassiou2024seed} is used to synthesize the command "Honey" for the command-based FDM. 
\paragraph{Voice-based FDM. }In this setting, LSLM can be interrupted by a variety of different words. The Speech Commands Dataset~\citep{zen2019libritts} is a set of one-second audio, each containing a single spoken English word. We split the dataset into training, validation, and test sets in an $8:1:1$ ratio, resulting in $51,088$, $6,798$, and $6,835$ pieces of data, respectively. In addition, we use a speaker independence setting, which guarantees that the speakers in the test set do not appear in the training set, simulating more challenging and realistic scenarios. 
\begin{table}[htbp]
    \centering
    \caption{Data details involved in training LSLM. SD means speaker dependence, while SI means speaker independence here. }
    \label{tab:data_details}
    \resizebox{\textwidth}{!}{
    \begin{tabular}{cc|cc}
        \hline
         & & \textbf{Command-based FDM(SD)} & \textbf{Voice-based FDM(SI)} \\
        \hline
        \multirow{3}{*}{TTS}
        & train      & \multicolumn{2}{c}{LibriTTS-train~\citep{zen2019libritts}} \\
        & val & \multicolumn{2}{c}{LibriTTS-dev-clean/other~\citep{zen2019libritts}} \\
        & test      & \multicolumn{2}{c}{LibriTTS-testsetB~\citep{du2024unicats}} \\
        \hline
        \multirow{4}{*}{Interruption} 
        & train & \multirow{3}{*}{Say\_Honey} & Speech Commands Dataset-train~\citep{warden2017speech} \\
        & val & & Speech Commands Dataset-dev~\citep{warden2017speech} \\
        & test & & Speech Commands Dataset-test~\citep{warden2017speech} \\
        \hline
        Noise & all        & \multicolumn{2}{c}{Freesound portion of MUSAN~\citep{snyder2015musan}} \\
        \hline
    \end{tabular}
    }
\end{table}

\subsection{Training and Inference Details}
We train the model with TTS, interruption, and noise datasets for $20$ epochs. 
For each sample, noise is added with a $50\%$ probability, and interruption with a $50\%$ probability, to the listening tokens. 
If a sample is selected to include an interruption, we modify the sentence to output the IRQ token $\mu = 0.5$ seconds after the start of the interruption and then stop outputting the remaining speaking tokens. 
This ensures that the model can correctly handle different audio signal combinations in the listening channel. 
The optimization strategy involves using AdamW~\citep{loshchilov2017decoupled} with a max learning rate of $5 \times 10^{-4}$ without weight decay and a batch size of $4$. 
The learning rate scheduler involves a warm-up phase for the first $5,000$ steps, followed by a cosine decay of the learning rate. 
Validation is performed at the end of each epoch, and the checkpoint with the lowest loss is selected for inference. 
The generation process employs Top-P sampling with a top-p value of $0.99$ and a temperature of $1.0$. 

\section{Experiments}
\subsection{Evaluation Metrics}

\paragraph{TTS capability evaluation.} We evaluate whether the speech generation capability is affected by the full duplex modeling in the proposed LSLM. 
The word error rate (WER) comparing the generated speech to the original text is considered as the TTS capability evaluation metrics using Whisper large v3\footnote{\url{https://github.com/openai/whisper}}~\citep{radford2023robust}. 
\paragraph{Interactive capability evaluation. } Interactivity capability evaluation aims to measure how well the proposed LSLM responds to real-time and unpredictable input from the listening channel. 
A successful turn-taking is defined as the model stopping speaking within the $[0, 2\mu]$ interval ($1$ second in our setting) after the interruption begins. 
Based on this, we categorize the outcomes into four cases: interruption and hit (TP), interruption and miss (FN), no interruption and hit (FP), and no interruption and miss (TN). 
From these cases, we construct a confusion matrix and calculate the Precision, Recall, and F1 score. 
These metrics consider both the success rate of turn-taking (Recall) and the rate of misjudgments (Precision), providing a comprehensive evaluation of the model's interactivity capabilities. 

\subsection{Experiments results}
We conduct a series of experiments to evaluate the command-based and voice-based FDM for both TTS capability and interactive capability. 
For TTS capability, we use a test set consisting of $500$ utterances, referred to as LibriTTS-testsetB~\citep{du2024unicats}, without any interruptions in the listening channel. The primary metric for this evaluation is WER. 
For the interactive capability evaluation, we employ a set of $1000$ utterances divided into two equal parts: $500$ utterances with interruptions at a random time step and $500$ utterances without interruptions. Interactive capability is measured using Precision, Recall, and F1 Score. 

Additionally, we test the models under two listening channel conditions: without noise, donated as Clean, and with noise, donated as Noise. For the baseline Vanilla TTS model, since it does not involve a listening channel, the input is inherently clean. By comparing the clean scenarios, we assess whether the intrinsic TTS capability is affected. Additionally, integrating noisy external inputs provides a better simulation of real-world scenarios. 

\paragraph{Command-based FDM.} 
For command-based FDM, we test the three architectures described in Section~\ref{sec:FDM-Ability} to fuse the listening channel and the speaking channel, which are early fusion (LSLM$_{EF}$), middle fusion (LSLM$_{MF}$), and late fusion (LSLM$_{LF}$). The results are shown in Table~\ref{tab:Command-based-FDM}. 
For TTS capability, The baseline Vanilla TTS model without a listening channel achieves a WER of $4.28\%$. 
LSLM$_{MF}$ outperforms LSLM$_{EF}$ and LSLM$_{LF}$ with a WER of $4.05\%$ in clean conditions and maintains a relatively low WER of $4.51\%$ in noisy conditions. 
The TTS ability of LSLM$_{EF}$ shows a notable decrease, likely due to the fusion of input embeddings,  making it difficult for the model to distinguish the information of the listening and speaking channels, negatively impacting the next token prediction. 
For interactive capability, all three architectures perform well with an oracle clean listening channel. However, LSLM$_{LF}$ shows a notable drop in performance under noisy conditions, with the F1 score falling to $94.89\%$. 
Observing that the late fusion method appears to mainly affect the precision score when the listening channel is noisy, suggests that the LSLM$_{LF}$ model reduces the discrimination of noise and human voice, leading to misjudgments of interruptions. 
In summary, the middle fusion approach demonstrates superior performance in TTS capability and competitive performance in interactive capability. Therefore, LSLM$_{MF}$ is concluded to be the best-performing model among those tested. 

\begin{table*}[htbp]
\centering
\caption{Experiments results on command-based FDM. Early fusion (LSLM$_{EF}$), middle fusion (LSLM$_{MF}$), and late fusion (LSLM$_{LF}$) are considered. }
\label{tab:Command-based-FDM}
\resizebox{1\textwidth}{!}{
\begin{tabular}{l|c|c|ccc}
\hline
\multirow{2}{*}{\textbf{Model}} & \multirow{2}{*}{\textbf{Listening Channel}} & \multicolumn{1}{c|}{\textbf{TTS Capability}} & \multicolumn{3}{c}{\textbf{Interactive Capability}} \\
& & WER(\%) $\downarrow$ & Precision(\%)$\uparrow$ & Recall(\%)$\uparrow$ &F1(\%)$\uparrow$ \\
\hline
Vanilla TTS & - (Clean) & 4.28 & - & - & - \\
\hline
\multirow{2}{*}{LSLM$_{EF}$} & Clean & 33.56 & 98.00 & 98.20 & 98.10 \\
 & Noise & 34.99 & 97.20 & 97.20 & 97.20 \\
\hline
\multirow{2}{*}{LSLM$_{MF}$} & Clean & 4.05 & 97.80 & 98.19 & 98.00 \\
 & Noise & 4.51 & 97.58 & 97.18 & 97.38 \\
\hline
\multirow{2}{*}{LSLM$_{LF}$} & Clean & 4.37 & 97.99 & 97.80 & 97.89 \\
 & Noise & 6.87 & 93.06 & 96.79 & 94.89 \\
\hline
\end{tabular}
}
\end{table*}

\paragraph{Voice-based FDM.} 
We utilized a more diverse set of interruption commands compared to the command-based FDM and involved unseen speakers in the testing procedures. The best configuration from the command-based FDM, the LSLM$_{MF}$ model, was selected to evaluate the voice-based FDM capability. The results are shown in Table~\ref{tab:Voice-based-FDM}. 
LSLM shows a higher WER of $5.33\%$ in clean conditions and $8.50\%$ in noisy conditions compared to the Vanilla TTS model, demonstrating the challenges posed by the real-world turn-taking problem. 
Comparing the results with the command-based FDM using the LSLM${_MF}$ model, we find that the voice-based setting faces greater challenges in maintaining high performance, especially under noisy conditions with Precision at $87.69\%$, Recall at $82.77\%$, and an F1 score of $85.15\%$. The diverse set of interruption commands and the involvement of unseen speakers add complexity, resulting in higher error rates. 

\begin{table*}[htbp]
\centering
\caption{Experiments results on voice-based FDM. LSLM here utilizes the architecture of middle fusion. }
\label{tab:Voice-based-FDM}
\resizebox{1\textwidth}{!}{
\begin{tabular}{l|c|c|ccc}
\hline
\multirow{2}{*}{\textbf{Model}} & \multirow{2}{*}{\textbf{Listening Channel}} & \multicolumn{1}{c|}{\textbf{TTS Capability}} & \multicolumn{3}{c}{\textbf{Interactive Capability}} \\
& & WER(\%) $\downarrow$ & Precision(\%)$\uparrow$ & Recall(\%)$\uparrow$ &F1(\%)$\uparrow$ \\
\hline
Vanilla TTS & - (Clean) & 4.28 & - & - & - \\
\hline
\multirow{2}{*}{LSLM } & Clean & 5.33 & 95.21 & 95.78 & 95.50 \\
& Noise & 8.50 & 87.69 & 82.77 & 85.15 \\
\hline
\end{tabular}
}
\end{table*}

\begin{figure*}[htbp]
  \centering
  \includegraphics[width=1\textwidth]{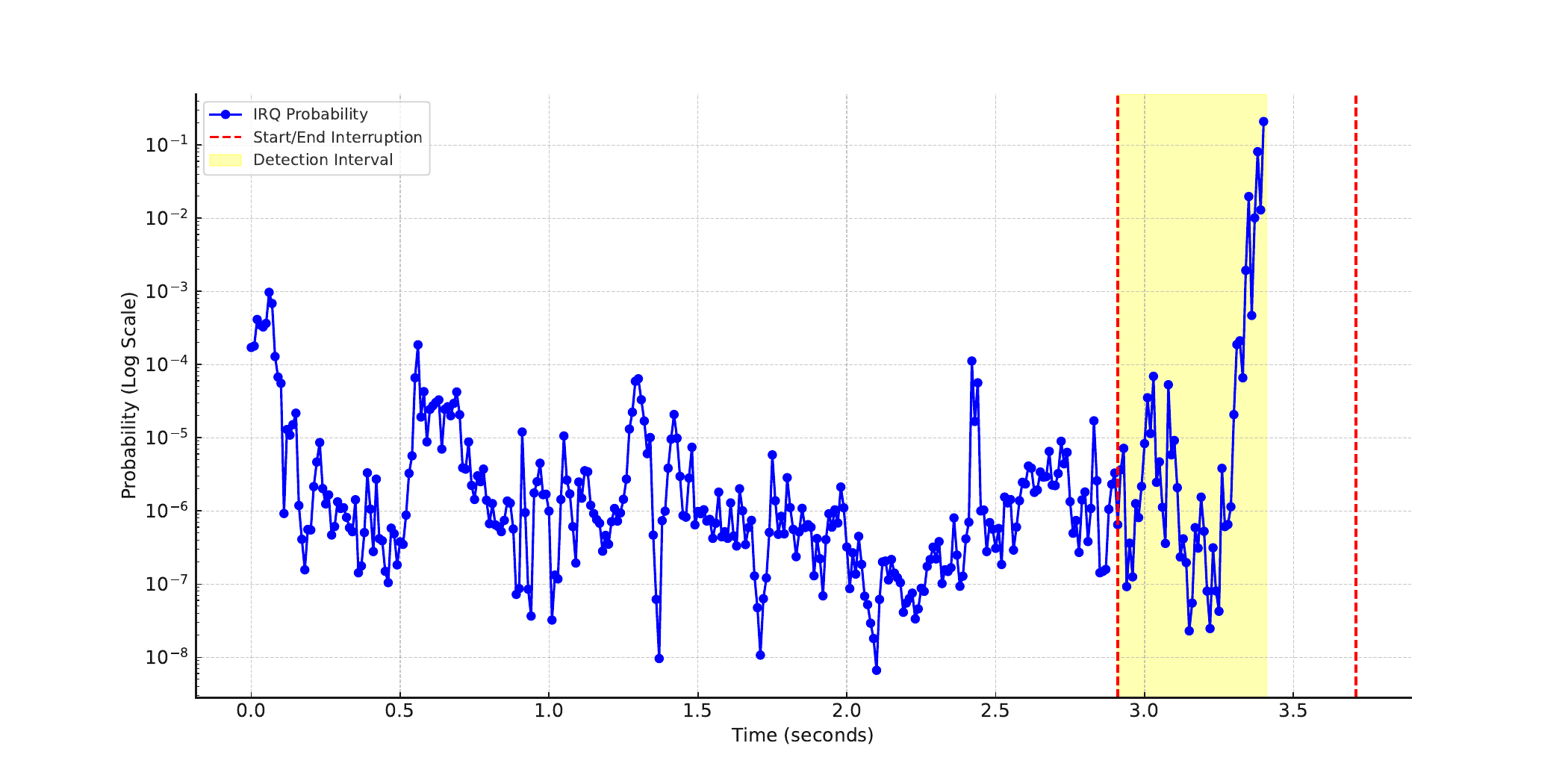}
  \caption{Illustration of the probability distribution of IRQ tokens (being interrupted) over time. The logarithmic scale probability is used for clear visualization. }
  \label{fig:visualization}
\end{figure*}

\paragraph{Visualization.} 
To investigate the turn-taking internal mechanism of LSLM, we visualize the probability distribution of IRQ tokens at different time steps during the generation process. 
Given that the IRQ token probability distribution varies significantly in order of magnitude across different time steps, we utilize a logarithmic scale for probability to enhance the clarity of the visualization. 
As illustrated in Figure~\ref{fig:visualization}, the probability of the IRQ token remains below $1\times10^{-3}$ when the model is not interrupted. 
When the listening channel starts to receive the real-time turn-taking signal, LSLM senses whether it is an interruption or a noise. 
After a very short time, the IRQ token probability begins to increase. 
Shortly thereafter, this probability rises to a level where the IRQ token is sampled by the model during generation. 

\subsection{Ablation Study}
In this section, we conduct an ablation study on LSLM with middle fusion architecture to evaluate the impact of different training methods on the performance of TTS capability and interactive capability. 
The training methods are categorized as training from scratch (\ding{55}), loading the pre-trained model and fixing the parameters (\ding{51}), and loading the pre-trained model and continuing training (\ding{58}). The detailed results are presented in Table~\ref{tab:ablation}. 

The vanilla TTS model, trained from scratch, achieves a WER of $4.28\%$ concerning TTS capability. 
For the interactive capability, the vanilla TTS model does not have a listening channel, hence no metrics are available. 
For the LSLM model, the best performance is observed when both the TTS backbone and streaming SSL encoder are loaded and continue training (\ding{58} \& \ding{58}), achieving the lowest WER of $4.05\%$ and highest Precision of $97.80\%$, Recall of $98.19\%$, and F1 Score of $98.00\%$. 
Some conclusions can also be drawn from these experiments. For example, the SSL encoder of the listening channel performs better when it can be continued training than fixed the parameters. One potential reason is that the SSL encoder has not encountered diverse noise during pre-training, creating a bottleneck for extracting audio with mixed human voice and noise when using fixed pre-trained parameters. 

\begin{table*}[htbp]
\centering
\caption{Ablation study on LSLM to evaluate the impact of different training methods. \ding{55} means training from scratch, \ding{51} means load the pre-training model and fix the parameters, \ding{58} means load the pre-training model and continue training. LSLM here utilizes the architecture of middle fusion. }
\label{tab:ablation}
\resizebox{1\textwidth}{!}{
\begin{tabular}{l|cc|c|ccc}
\hline
\multirow{2}{*}{\textbf{Model}} & \multicolumn{2}{c|}{\textbf{Training Method}} & \multicolumn{1}{c|}{\textbf{TTS Capability}} & \multicolumn{3}{c}{\textbf{Interactive Capability}} \\
& Speaking & Listening & WER(\%) $\downarrow$ & Precision(\%)$\uparrow$ & Recall(\%)$\uparrow$ &F1(\%)$\uparrow$ \\
\hline
Vanilla TTS & \ding{55} & - & 4.28 & - & - & - \\
\hline
\multirow{5}{*}{LSLM} & \ding{55} & \ding{51} & 4.82 & 97.80 & 97.99 & 97.89 \\
 & \ding{55} & \ding{58} & 4.67 & 95.60 & 95.98 & 95.79\\
 & \ding{51} & \ding{51} & 6.64 & 97.89 & 83.60 & 90.18 \\
 & \ding{51} & \ding{58} & 4.64 & 97.60 & 98.18 & 97.89 \\
 & \ding{58} & \ding{51} & 4.46 & 96.43 & 92.54 & 94.44\\
 & \ding{58} & \ding{58} & 4.05 & 97.80 & 98.19 & 98.00 \\
\hline
\end{tabular}
}
\end{table*}

\section{Conclusion}

In this paper, we address the challenges of enhancing real-time interaction by introducing full duplex modeling (FDM) in interactive speech language models (iSLM). 
We introduce listen-while-speaking language model(LSLM), an innovative end-to-end model designed to handle real-time turn-taking. LSLM integrates a token-based decoder-only TTS model for speech generation and a streaming SSL encoder for audio input, enabling simultaneous listening and speaking. 
We propose three strategies for fusing duplex signals: early fusion, middle fusion, and late fusion. Among these, Middle Fusion demonstrates a superior balance between speech generation and real-time interaction capabilities. 
The proposed LSLM is evaluated in two settings: command-based FDM and voice-based FDM. Our experiments show that LSLM is robust to noisy environments and responsive to diverse instructions from unseen speakers, achieving effective duplex communication with minimal impact on system performance. 
Our work is an initial exploration into full duplex interactive speech language models, and there is still a long way to go to achieve smooth human-computer speech interaction. 
There is a lot to explore in the future, such as developing speech-in speech-out dialogue systems with full duplex modeling ability, incorporating speaker-following capability to identify interrupting speakers, and exploring audiovisual co-guidance for improved turn-taking. 

\bibliographystyle{plainnat}
\bibliography{custom}

\end{document}